\documentclass{article}
\usepackage[final]{corl_2026}

\usepackage{graphicx}
\usepackage{booktabs}
\usepackage{amsfonts}
\usepackage{amsmath}
\usepackage{amssymb}
\usepackage{url}

\title{When Does Dynamics Randomization Test Context Inference?}

\author{%
    Oleg Smirnov$^{1}$ \And
    Sofiane Ennadir$^{1}$ \And
    John Pertoft$^{1}$ \And
    Bjartur Hjaltason$^{1}$ \And
    Sara Karimi\rlap{$^{1,2}$}%
}

\begin{document}
\maketitle

\footnotetext[1]{King AI Labs, Microsoft Gaming. Corresponding author: \texttt{oleg.smirnov@microsoft.com}}
\footnotetext[2]{KTH Royal Institute of Technology}

\begin{abstract}
Environments with randomized dynamics are widely used to ask whether world
models need explicit context inference. However, randomization can leave
little predictive information for a context encoder to add, because distinct
latent parameters may produce similar transitions at visited states, or those
states may already reveal the active context. We formalize the remaining
one-step information as \emph{predictive headroom} and compare kernels under
fixed and visited state--action distributions to distinguish changes in query
coverage from changes in context weights. We analyze discrete and continuous
environments and find low headroom both when kernels differ little and when
visitation reduces an otherwise large discrepancy. At the same time, resetting
state components increases headroom, showing that context evidence accumulates
in state. We propose reporting this headroom alongside the randomized
parameters, so that gains from explicit context inference can be interpreted
relative to the information missing from the current state and action.
\end{abstract}

\section{Introduction}
\label{sec:introduction}
In model-based RL, environments with randomized dynamics are commonly used to
test whether world models benefit from explicit context inference. Latent
physical parameters, such as masses and forces, vary across episodes in
CARL~\citep{benjamins2022contextualize} and the Real-World RL
suite~\citep{dulacarnold2021challenges}, and context-conditioned world
models~\citep{prasanna2024dreaming,roder2025dynamics} are compared against
context-blind predictors under such variation. However, a benchmark can
randomize parameters and still leave a context mechanism little predictive
information to add. Consider a grid world with a hidden wind that displaces
the agent toward one wall after each move. An agent crossing the open interior
drifts downwind, so its position carries evidence about the wind. An agent
moving along the downwind wall shows no drift, because the wall blocks it. In
neither case does revealing the wind add much to one-step prediction.

The example shows the two mechanisms that can reduce the remaining predictive
information. First, kernels that differ substantially across the state space
may differ little at the visited state--action pairs, as at the wall. Second,
the dynamics can make visited states informative about the hidden parameter.
Repeated displacement by the wind, or the change a persistent force makes to
position and velocity, leaves evidence about the parameter in the current
state. Even a memoryless model can then use this evidence without a history
encoder. A small gain from revealing the true context can thus reflect weak
kernel differences or implicit identification.

Contextual MDPs formalize latent dynamics that vary across
episodes~\citep{hallak2015contextual,doshi2016hidden}. Online system
identification conditions a policy on parameters inferred from recent
experience~\citep{yu2017preparing}. Adaptive dynamics
models~\citep{nagabandi2019learning} and context-aware
predictors~\citep{lee2020context,evans2022context} use experience to improve
prediction under changing parameters. While those methods make predictive
structure explicit, we quantify how much information about transitions the
current state withholds before any such mechanism is added.

We introduce an experimental diagnostic of context information under a given
visitation. We compare transition kernels under a fixed reference and at
visited state--action pairs using prior context weights, then reweight
contexts by visitation to estimate the remaining predictive headroom. We show,
in discrete and continuous settings, that similar headroom can arise from
differing kernel variation. We propose reporting this measure when comparing
world models, because explicit context inference can add predictive
information only where the inputs miss it.

\section{Predictive headroom} \label{sec:comparisons}
Consider a contextual MDP with observed state $s$, action $a$, and transition
kernel $T_\theta(s'\mid s,a)$. Context $\theta\sim p(\theta)$ is drawn
independently at episode start and held fixed. Rewards are unchanged across
contexts and are not used by the diagnostic. Let $d(\theta,s,a)$ be the
transition-weighted joint distribution produced by a specified collection
procedure: pooled transitions receive equal weight. Its marginal $d(s,a)$
describes the visited queries, and $d(\theta\mid s,a)$ describes which
contexts remain plausible at each query.

\textbf{Predictive headroom.} The optimal predictor uses the mixture $\bar
T_d(\cdot\mid s,a)= \mathbb{E}_{\theta\sim d(\theta\mid
  s,a)}T_\theta(\cdot\mid s,a)$, whereas revealing context instead gives
$T_\theta$. Their minimum expected one-step log-loss difference is
\begin{equation}
\mathcal{I}_{\mathrm{emp}}
=\mathbb{E}_{(\theta,s,a)\sim d}
 D_{\mathrm{KL}}\!\left(T_\theta(\cdot\mid s,a)
 \,\middle\|\,\bar T_d(\cdot\mid s,a)\right)
=I_d(\theta;s'\mid s,a).
\label{eq:headroom}
\end{equation}
This quantity measures the transition information context adds beyond the
current $(s, a)$ pair, and can be estimated using known kernels and fitted
context mixtures. While the target concerns information in the inputs, a
learned world model can still incur approximation or optimization error.

\textbf{Fixed-reference and matched-query variation.} Define the prior mixture
$\bar T_p(\cdot\mid s,a)= \mathbb{E}_{\theta\sim p(\theta)}T_\theta(\cdot\mid
s,a)$ and the local prior-weighted separation $J(s,a)=\mathbb{E}_{\theta\sim
  p(\theta)} D_{\mathrm{KL}}(T_\theta(\cdot\mid s,a)\| \bar T_p(\cdot\mid
s,a))$. We compare
\begin{equation}
\mathcal{I}_{\mathrm{struct}} =\mathbb{E}_{(s,a)\sim\mu}J(s,a), \qquad
\mathcal{I}_{\mathrm{match}} =\mathbb{E}_{(s,a)\sim d(s,a)}J(s,a),
\label{eq:references}
\end{equation}
where $\mu$ is a fixed, context-independent query distribution. Therefore,
$\mathcal{I}_{\mathrm{struct}}$ can be used to compare kernels at common
queries, while $\mathcal{I}_{\mathrm{match}}$ is used to evaluate their
differences where data were collected, retaining prior context weights in
both. In the wind example, $\mathcal{I}_{\mathrm{struct}}$ compares the
kernels at every interior state--action pair with every wind equally
weighted, $\mathcal{I}_{\mathrm{match}}$ compares them at the pairs the agent
visits, weighted by visit frequency, with the same wind weights, and
$\mathcal{I}_{\mathrm{emp}}$ compares them at those pairs with each wind
weighted by how often the agent reaches them under that wind.
Table~\ref{tab:diagnostic-summary} lists the queries and context weights of
each quantity and reports the three values for this example and two others.

The structural-to-matched comparison changes query coverage only. The
matched-to-visited comparison holds queries fixed and replaces prior context
weights by $d(\theta\mid s,a)$ in both the outer average and the predictive
mixture. This reweighting reflects state-dependent evidence and any marginal
shift caused by pooling episodes of unequal length. We note that since the
comparisons use different distributions, those quantities do not constitute
general bounds on one another.

Low matched variation indicates that context changes little about one-step
outcomes on the visited support. A large matched-to-visited reduction shows
that context reweighting matters. To attribute that reduction to
state-mediated identification, we must account for changes in marginal context
weights $d(\theta)$. A concentrated marginal alone can reduce headroom even
when state is independent of context.

\section{Controlled evidence} \label{sec:evidence}
\textbf{Environments.} MiniGrid~\citep{chevalier2023minigrid} provides an
enumerable state space $(x,y,d)$, with position and direction on an
\texttt{Empty-8x8} grid. We use a \emph{Wind} variation that samples
independent drift coordinates from $\mathcal{U}[-v_{\max},v_{\max}]$. After
each native action, each axis moves one cell in the sampled direction with
probability equal to its drift magnitude, subject to wall blocking. In another
variation, \emph{Resistance} samples an action-failure probability
$\epsilon\sim\mathcal{U}[0,\epsilon_{\max}]$, and a failed action leaves the
state unchanged. Finally, we consider \emph{Point-Mass-Attractor} as a
continuous two-dimensional system with observed position and velocity, where
one of four hidden corners $(\pm1.5,\pm1.5)$ exerts force $-k(\mathbf
p-\theta)$ alongside the action, with independent position and velocity noise
and reflecting boundaries.

We experiment with $5{,}000$ episodes in each setting under uniform random
actions: up to $256$ steps in MiniGrid and $50$ in Attractor, where the fixed
episode length avoids duration-induced context reweighting. Context labels are
logged for the diagnostic and excluded from the state--action predictive
information set. We evaluate transition prediction without training policies
or comparing returns.

\textbf{Estimation.} MiniGrid's structural reference is uniform over interior
positions, four directions, and seven actions. There, we enumerate the kernels
and integrate the continuous context prior. Visited headroom averages exact
kernel KL divergences against mixtures fitted at each $(s, a)$ pair, leaving
out the scored episode and applying one pseudo-count of smoothing. In
Attractor, we use its query-invariant interior Gaussian kernel for the prior
references. Its visited estimate scores logged next states under the known
kernels, including boundary reflection, against a mixture with a five-fold,
episode-split nearest-neighbor context posterior.

The estimates retain all transitions. MiniGrid refits its mixtures, whereas
Attractor resamples fixed per-episode contributions. We report
fitted-predictor estimates with bootstrap uncertainties
(Section~\ref{sec:limitations}). Implementation details are deferred to
Appendix~\ref{app:details}.

\begin{figure}[t]
\centering
\includegraphics[width=\linewidth]{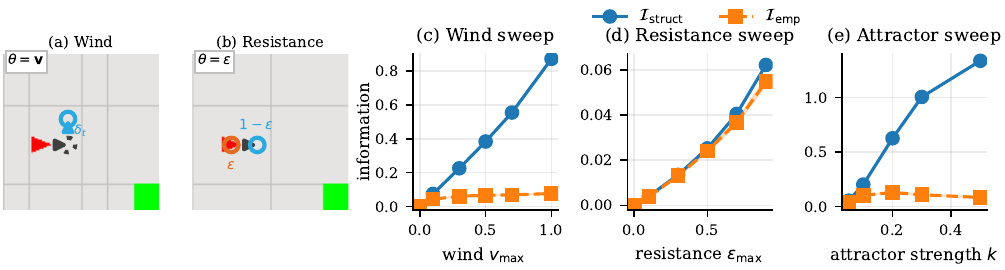}
\caption{Dynamics variation and predictive headroom across perturbation
  strengths. (a,b) Wind and Resistance mechanisms. (c--e) The structural
  reference $\mathcal{I}_{\mathrm{struct}}$ and estimated visited headroom
  $\widehat{\mathcal{I}}_{\mathrm{emp}}$ differ substantially on Wind and
  Attractor but remain close on Resistance. Attractor's structural curve uses
  the interior Gaussian reference. Shading shows bootstrap intervals.}
\label{fig:diagnostic-sweeps}
\end{figure}

\begin{table}[t]
\centering
\footnotesize
\caption{Comparisons at representative settings. The first two rows give the
  query distribution and the context weights of each quantity. Attractor's
  identical prior references use its query-invariant interior kernel.
  Uncertainties are the larger one-sided deviations from the $95\%$ bootstrap
  intervals. }
\label{tab:diagnostic-summary}
\begin{tabular}{@{}lccc@{}}
\toprule
 & Fixed & Matched & Visited \\
 & $\mathcal{I}_{\mathrm{struct}}$ & $\mathcal{I}_{\mathrm{match}}$
 & $\widehat{\mathcal{I}}_{\mathrm{emp}}$ \\
\midrule
Queries & $\mu$ & $d(s,a)$ & $d(s,a)$ \\
Context weights & $p(\theta)$ & $p(\theta)$ & $d(\theta\mid s,a)$ \\
\midrule
Wind ($v_{\max}=1$) & 0.871 & 0.620 & $0.077 \pm 0.002$ \\
Resistance ($\epsilon_{\max}=0.9$) & 0.062 & 0.059 & $0.055 \pm 0.001$ \\
Attractor ($k=0.3$) & 1.006 & 1.006 & $0.109 \pm 0.002$ \\
\bottomrule
\end{tabular}
\end{table}

\textbf{Sources of low headroom.} Figure~\ref{fig:diagnostic-sweeps}
illustrates that visited headroom can remain low as dynamics variation
increases. In Wind (Table~\ref{tab:diagnostic-summary}), matching the query
distribution reduces the reference by $29\%$, so coverage accounts for part of
the difference. Using visitation-dependent context weights at those same
queries removes a further $88\%$ of the matched reference.

In Resistance, prior-weighted separation changes little under either
comparison, and visited headroom retains about $93\%$ of the matched
reference. Its low headroom mostly reflects weak kernel differences under
these query and action weights. In Attractor, the prior reference is large but
estimated visited headroom is about $11\%$ of that interior reference.

\begin{table}[t]
\centering
\footnotesize
\caption{State resets increase predictive headroom without changing episode
  context. Full resets resample position and direction in Wind. In Attractor
  they resample position and zero velocity. Changes in parentheses use each
  intervention dataset's own no-reset baseline, collected separately from the
  sweep datasets.}
\label{tab:resets}
\begin{tabular}{@{}lcc@{}}
\toprule
Reset & Wind & Attractor \\
\midrule
None & $0.078 \pm 0.002$ & $0.112 \pm 0.002$ \\
Full, every 10 steps & $0.411 \pm 0.006$ $(+430\%)$ & $0.473 \pm 0.004$ $(+324\%)$ \\
Full, every 2 steps & $0.723 \pm 0.017$ $(+833\%)$ & --- \\
Position only, every 10 steps & --- & $0.143 \pm 0.003$ $(+28\%)$ \\
\bottomrule
\end{tabular}
\end{table}

\textbf{State-component resets.} We reset selected components before recording
the current state, keeping the episode context and uniform action rule
unchanged. Wind resamples valid position and direction uniformly, while
Attractor resamples position uniformly in $[-2,2]^2$ and either zeroes or
preserves velocity. Table~\ref{tab:resets} shows that frequent Wind resets
increase predictive headroom. In Attractor, resetting both position and
velocity raises headroom more than resetting position alone. The resets leave
the transition family fixed while changing both query coverage and the
posterior over context. They show that context information accumulates in
state, including velocity, but do not isolate its effect from changes in
coverage.

\section{Implications for evaluating world models} \label{sec:implications}
Low headroom can explain a small gain from explicit context inference when the
observed state already supplies most of the evidence needed for prediction.
High headroom instead indicates that revealing context can improve
prediction. However, one should separately measure which information a learned
representation retains and how well its predictor uses it, since a context
encoder can still improve learning or approximation even when headroom is
low. Comparisons of predictors with and without explicit context inference
should therefore measure both the information available in the inputs and the
errors of the actual models under matched data and compute.

We propose reporting whether a benchmark's context-conditional kernels remain
predictively distinct at overlapping observed states. This requires specifying
the collection rule, predictive inputs, transition weighting, and marginal
context frequencies $d(\theta)$ relative to the design prior. A low matched
reference suggests changing the queries or perturbations, whereas a large
matched-to-visited reduction requires distinguishing context imbalance from
information retained in state. Increasing the parameter range alone does not
resolve either issue reliably. We can compute these comparisons from context
labels and simulator kernels, but with learned kernels we must also account
for prediction error.

Initial-state distributions, episode length, and observed physical coordinates
also influence how much context remains hidden. A reset that changes position
but preserves velocity can retain context information that the benchmark is
intended to withhold. We therefore recommend reporting matched-query variation
under each reset protocol's collection distribution to separate changes in the
prediction support from changes in context weights when interpreting the
resulting headroom. We propose examining the evaluation support before
interpreting task success or a small context benefit as evidence about the
need for explicit predictive structure.

\section{Conclusion} \label{sec:conclusion}
We posit that dynamics randomization alone does not establish that a benchmark tests context
inference. Our controlled examples show that similar predictive headroom can
accompany very different kernel variation, and that observed state can retain
evidence about hidden dynamics. Reporting this headroom would help distinguish
limited opportunity for context inference from weaknesses of the methods being
compared, making such benchmarks more informative about when explicit context
inference in world models has information to add.

\section*{Limitations} \label{sec:limitations}
The three environments used in the paper are synthetic benchmarks with known
transition kernels. All data was collected under uniformly random actions, so
any context evidence in the predictor inputs comes from state. A trained
policy changes $d(s,a)$ and $d(\theta\mid s,a)$, and headroom under that
policy has to be estimated from its own data. Moreover, the reported values
are predictor estimates rather than exact mutual information.

Our measurements concern one-step prediction on the collection distribution,
while a model deployed for planning or new goals may need predictions at
states and actions which are rarely represented there. As such, for learned
policies or physical robots, the diagnostic requires measurements under their
own prediction and control conditions.

\section*{AI use statement}
Generative AI tools were used to assist with software implementation,
including environment and analysis scripts, and to edit the manuscript for
language, wording, and phrasing. The authors reviewed all code, analyses, and
text and take full responsibility for the final content of this work.

\bibliography{main}

\clearpage
\appendix

\section{Implementation and estimation details} \label{app:details}
\begin{figure}[b]
\centering
\includegraphics[width=\textwidth]{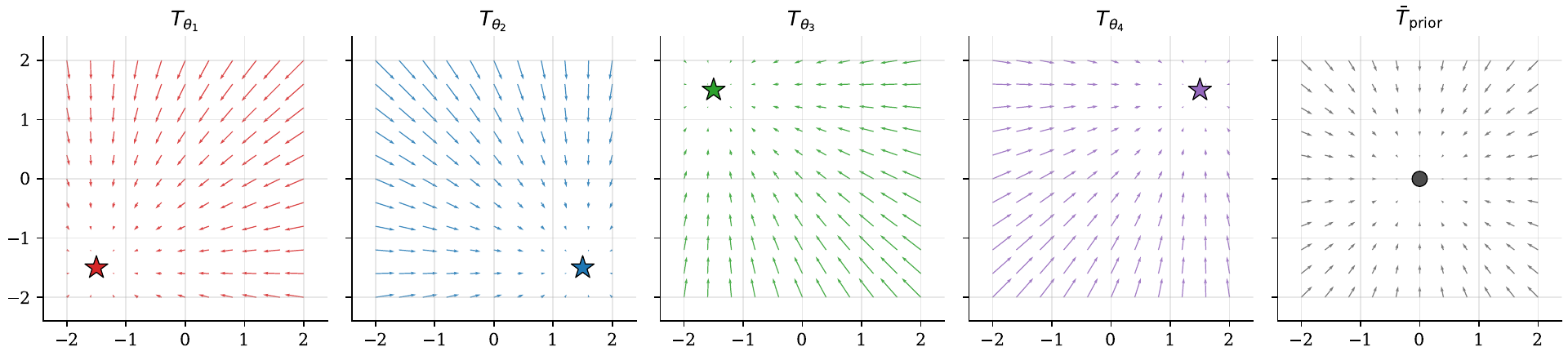}
\caption{Point-Mass-Attractor force fields. The first four panels show the context-dependent force $-k(\mathbf p-\theta_i)$ for each kernel $T_{\theta_i}$. The final panel shows the prior-averaged force $-k\mathbf p$.}
\label{fig:attractor_force_field}
\end{figure}

\textbf{MiniGrid kernels.} Both families retain the \texttt{Empty-8x8} layout,
goal, and reward. The logged episodes start at $(x,y,d)=(1,1,0)$, facing
right, and end on goal termination at $(6,6)$ or truncation at $256$ steps.
The native action updates position or direction first. For Wind, each
coordinate $v_j$ specifies drift $\operatorname{sign}(v_j)$ with probability
$|v_j|$, and no drift otherwise. A proposed displacement into a wall is
blocked.  Resistance executes the native action with probability $1-\epsilon$
and otherwise leaves position and direction unchanged. The fixed reference
enumerates $6\times6\times4\times7=1008$ state--action queries. Continuous
contexts are integrated by quadrature, without context binning. The sweeps use
$v_{\max}\in\{0,0.1,0.3,0.5,0.7,1.0\}$ and
$\epsilon_{\max}\in\{0,0.1,0.3,0.5,0.7,0.9\}$.

For visited headroom, we group known kernels at logged contexts by
state--action cell. For each scored episode, we fit the mixture using the
other episodes and add one uniform pseudo-row over that cell's kernel outcome
support.  The score integrates KL over the discrete next-state outcomes rather
than using only the realized next state. The $200$ percentile bootstrap
resamples reweight entire episodes and recompute the mixtures. The point
estimate assigns equal weight to pooled transitions.

\textbf{Attractor dynamics.} The point mass has unit mass,
timestep $\Delta t=0.05$, and uniformly sampled context
$\theta\in\{(\pm1.5,\pm1.5)\}$ (Figure~\ref{fig:attractor_force_field}). With force $\mathbf f=\mathbf a-k(\mathbf
p-\theta)$ and independent actions $\mathbf a_t\sim\mathcal{U}[-1,1]^2$, the
noiseless update is $\mathbf v^+=\mathbf v+\Delta t\,\mathbf f$ followed by
$\mathbf p^+=\mathbf p+\Delta t\,\mathbf v^+$.  Independent Gaussian
increments have per-coordinate variances $\sigma^2\Delta t$ for velocity and
$\sigma^2\Delta t^3$ for position, with $\sigma=0.1$. A single reflection per
axis at the workspace boundaries $\pm2.5$ reverses velocity with damping
factor $0.5$.  Initial position is uniform in $[-0.5,0.5]^2$ and initial
velocity is zero.  The sweep uses $k\in\{0.05,0.1,0.2,0.3,0.5\}$.

Before reflection, context changes only the Gaussian mean displacement, so the
prior-weighted information is independent of the query. We evaluate both fixed
and matched references from this interior kernel by one-dimensional quadrature for each coordinate. For visited scores, we compute
the reflected density by summing its preimages with their Jacobians.

\textbf{Attractor posterior fitting.} Five episode-level
folds produce nearest-neighbor class frequencies from standardized position
and velocity, with one Laplace pseudo-count for each of the four
contexts. Actions are independent uniform draws, so the posterior uses state
features only. Neighbor counts in $\{128,256,512,1024\}$ are compared by
context log loss on the same scored folds. At each logged next state, the
estimator subtracts the fitted mixture log density from the true-context log
density and averages over all transitions.  The $1000$ percentile bootstrap
resamples use fixed episode sums and counts, without refitting or repeating
neighbor selection. These intervals summarize sampling variation conditional
on the fitted predictions, not the full estimation uncertainty. Even known
component kernels do not make the visitation-dependent posterior exact.

As a local check, immediately after a zero-velocity position reset the context
posterior is the known uniform prior. At reset interval ten, the fitted mean
log-density contrast on those transitions is $1.0110$ nats, versus $0.9961$
when using the known prior. At interval two the corresponding values are
$1.0093$ and $1.0071$. This checks the estimator on the reset stratum, not on
all visited states.

\end{document}